\documentclass[sigconf]{acmart}
\AtBeginDocument{%
  \providecommand\BibTeX{{%
    \normalfont B\kern-0.5em{\scshape i\kern-0.25em b}\kern-0.8em\TeX}}}

\copyrightyear{2020} 
\acmYear{2020} 
\setcopyright{acmlicensed}
\acmConference[ICSE-NIER'20]{New Ideas and Emerging Results}{May 23--29, 2020}{Seoul, Republic of Korea}
\acmBooktitle{New Ideas and Emerging Results (ICSE-NIER'20), May 23--29, 2020, Seoul, Republic of Korea}
\acmPrice{15.00}
\acmDOI{10.1145/3377816.3381734}
\acmISBN{978-1-4503-7126-1/20/05}
\usepackage{algorithm,algorithmic}
\usepackage{subfig}
\usepackage{multirow}

\usepackage[super]{nth}
\usepackage{esvect}
\usepackage[skip=2pt]{caption}
\begin{document}

%%
%% The "title" command has an optional parameter,
%% allowing the author to define a "short title" to be used in page headers.
\title{Manifold for Machine Learning Assurance}

%%
%% The "author" command and its associated commands are used to define
%% the authors and their affiliations.
%% Of note is the shared affiliation of the first two authors, and the
%% "authornote" and "authornotemark" commands
%% used to denote shared contribution to the research.
\author{Taejoon Byun}
\orcid{1234-5678-9012}
\affiliation{%
  \institution{University of Minnesota}
  \streetaddress{200 Union St.}
  \city{Minneapolis}
  \state{Minnesota}
%  \postcode{55116}
}
\email{taejoon@umn.edu}

\author{Sanjai Rayadurgam}
\affiliation{%
  \institution{University of Minnesota}
  \streetaddress{200 Union St.}
  \city{Minneapolis}
  \state{Minnesota}
  %\city{Minneapolis}
  %\state{Minnesota}
%  \postcode{55116}
}
\email{rsanjai@umn.edu}

%%
%% By default, the full list of authors will be used in the page
%% headers. Often, this list is too long, and will overlap
%% other information printed in the page headers. This command allows
%% the author to define a more concise list
%% of authors' names for this purpose.
%\renewcommand{\shortauthors}{Trovato and Tobin, et al.}

%%
%% The abstract is a short summary of the work to be presented in the
%% article.
\begin{abstract}
%ML components have to be tested well.
%
The increasing use of machine-learning (ML) enabled systems in critical tasks 
fuels the quest for novel verification and validation techniques yet grounded 
in accepted system assurance principles.
In traditional system development, model-based techniques have been widely adopted, where the central premise is that abstract models of the required system provide a sound basis for judging its implementation. 
We posit an analogous approach for ML systems using an ML technique that extracts from the high-dimensional training data implicitly describing the required system, a low-dimensional underlying structure---a \textit{manifold}.
It is then harnessed for a range of quality assurance tasks such as test adequacy measurement, test input generation, and runtime monitoring of the target ML system.
%As most of the ML quality metrics (such as accuracy) base their measurement on the test data, it is important to make sure that we have a {\it good} test dataset in order to objectively assess the quality of the %learning-enabled systems.
%
%The ideas from the SE community is actively being transferred for better testing ML-enabled software
%
The approach is built on variational autoencoder, an unsupervised method for learning a pair of mutually near-inverse functions between a given high-dimensional dataset and a low-dimensional representation.
% for the data in a low-dimensional space
%learning a hidden low-dimensional structure from  a given dataset, along with a pair of functions that maps 
%the input dimension to and from the manifold dimension.
%
Preliminary experiments establish that the proposed manifold-based approach, for test adequacy drives diversity in test data, for test generation yields fault-revealing yet realistic test cases, and for run-time monitoring provides an independent means to assess trustability of the target system's output.

\end{abstract}

%%
%% The code below is generated by the tool at http://dl.acm.org/ccs.cfm.
%% Please copy and paste the code instead of the example below.
%%


%%
%% Keywords. The author(s) should pick words that accurately describe
%% the work being presented. Separate the keywords with commas.

%% A "teaser" image appears between the author and affiliation
%% information and the body of the document, and typically spans the
%% page.
%\begin{teaserfigure}
%  \caption{Seattle Mariners at Spring Training, 2010.}
%  \Description{Enjoying the baseball game from the third-base
%  seats. Ichiro Suzuki preparing to bat.}
%  \label{fig:teaser}
%\end{teaserfigure}

%%
%% This command processes the author and affiliation and title
%% information and builds the first part of the formatted document.
\maketitle

\section{Introduction}

Machine-learning enabled systems are being increasingly adopted in safety-critical applications. 
This brings into renewed focus the important problem of assurance of such systems and drives
research into novel and effective techniques for V\&V of ML systems.
%The increasing use of machine-learning (ML) components in safety-critical applications such as autonomous driving is driving the focus of software engineering researchers towards important questions of verification and validation of such systems~\cite{}.
%
%In particular, testing is adopted most widely in practice for assessing the ML system for various qualities such as correctness, robustness, fairness, and efficiency~\cite{zhang2019machine}.

In traditional development, model-based techniques are well-established and central to a variety of assurance activities. 
A model, by definition, abstracts away irrelevant system details, retaining only what is essential
for its intended purposes.
ML techniques produce such a {\it model} for a given task by learning the essential features from provided data.
%
%In ML, a key assumption is that there are implicit patterns to be gleaned
%from the provided data that sufficiently describe the desired system behavior.
%
%ML techniques help learn a good representation---\emph{a model}---that exhibits those patterns.
%
If a model useful for a task can be learned instead of being hand-coded, can models useful for quality assurance---such as behavioral model in traditional software engineering---also be similarly learned?
We answer this question affirmatively here using a specific learning technique that appears to be a good fit for multiple assurance activities.

Two considerations influence our choice of the learning approach for such assurance models:
The ability to (i) identify inputs of interest for exercising the target system; and
(ii) flag outputs of concern when the target system is exercised.
The former is useful for constructing various scenarios to assess the ML system, and the latter for judging its exhibited behavior, both very important for assurance tasks.
We investigated variational autoencoders, which learn to-and-fro maps between a high-dimensional input space and a low-dimensional manifold space---the learned implicit model of the data---such that their composition approximates the identity function on the input used for training.
%
%where the manifold---the learned implicit model of the data---resides,
%
If we can carve out regions of interest in this low-dimensional manifold space (the model domain), we can then use the two learned maps to effectively address both considerations.
This is the basis for the proposed manifold-based assurance techniques elaborated in the sequel.

In a set of preliminary experiments performed on image classification models, we find that the manifold-based coverage has a stronger correlation than neuron coverage to semantic features in the data. 
We also present results from manifold-based test generation which shows that realistic yet fault-revealing test inputs can be generated in volume, which can help in addressing a target system's weaknesses.
Finally, we briefly discuss a manifold-based runtime monitor to assess output trustability for a prototype ML-based aircraft taxiing system.

\section{Preliminaries}

% ----- Simple VAE Section ----- %

%Consider a large dataset of images such as  MNIST~\cite{cohen2017emnist} for hand-written digits.
%
%Suppose one can describe each individual image with a small number of "latent" variables (implicitly
%inferred), where each variable represents some relevant attribute that describes the image  such as, the digit it represents, its orientation, thickness, size, font family, and so on.
%
%If one can identify a complete set of variables with which to describe every image in the whole dataset, one can model similar images that are not present in the dataset, using similar patterns of combinations of values for these variables. 
%
%Furthermore, if an automated technique exists which can {\it learn} those variables automatically from the data set, and even generate images from the encoding,
%we can build an image generator that affords us some control over generation.
%
%There exists such technique, and it is called Variational Autoencoder (VAE).

%\textbf{Do we need subsection titles, or just paragraphs}
%\subsection{Manifold}
Mathematically, a \emph{manifold} is a topological space that is locally Euclidean (e.g. the surface of the Earth).
%In mathematics, manifold is defined as a topological space that locally resembles Euclidean space near each point.
%
In ML, manifold hypothesis states that real world data $X$ presented in high-dimensional spaces $R^{d_X}$ are expected to concentrate in the vicinity of a manifold $M$ of a much lower dimension $d_M$ embedded in $R^{d_X}$~\cite{bengio2013representation}.
%
%In other words, manifold hypothesis posits that high dimensional data---such as image---can be explained with a number of factors that is much smaller than the dimensionality of the input space.
%
Manifold learning tries to capture this mapping so that a complex dataset can be encoded into a meaningful representation in a smaller dimension, serving several purposes such as data compression and visualization~\cite{cayton2005algorithms}.
Among many manifold learning approaches, we built our techniques on top of variational autoencoder as they provide unique advantages over other methods, such as the capability of synthesizing new inputs with expected outputs from the manifold by interpolating among existing data points.

%{\bf Mention this work also}~\cite{lin2008reimannian}.

\begin{figure}[]
    \centering
    \includegraphics[width=0.85\columnwidth]{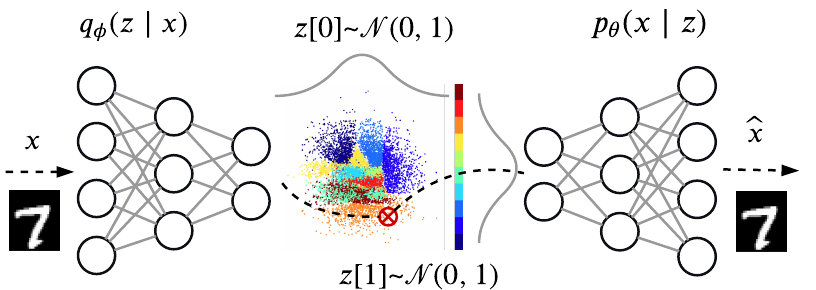}
    \caption{
    A VAE with $\kappa=2$.
    The colored points comprise a 2D manifold of MNIST encoded in $\mathcal{N}([0,0]^T, I_2)$.
    The color represents class labels from digit 0 (dark navy) to 9 (burgundy). }
    \label{fig:vae}
\end{figure}

%\subsection{Variational Autoencoder}
%\label{sec:vae}

Variational autoencoder (VAE) is a latent-variable generative model which is capable of producing outputs similar to inputs by determining a latent-variable space $Z$ and an associated probability density function (PDF) $P(z)$.
%sampled from the manifold $\mathcal{X}$ by determining a latent-variable space $\mathcal{Z}$ and associated density function $P(z)$.
%
Its goal is to ensure that for each datapoint $x$ in a given dataset $X$, there is at least one valuation of the latent variables $z \in Z$ which causes the model to generate $\hat{x}$ that 
is very similar to $x$.
This is achieved by optimizing $\theta$ for a deterministic function $f: Z \times \Theta \to X$ such that the random variable $f(z; \theta)$ produces outputs similar to $x \in X$ when $z$ is sampled from $P(z)$.
%
%In other words, we maximize the likelihood of producing $X$ when $X$ is conditioned by $Z$: $P(X) = \int P(X | z; \theta) P(z) dz$; here, a PDF $P(X|z; \theta)$ replaces $f(z; \theta)$.
%
VAE assumes that any probability distribution $P(z)$ can be obtained by applying a (sufficiently sophisticated) function $f_\theta$ to a set of independent normally distributed variables $z$.
%
%With a set of decoder parameters $\theta$, the probabilistic decoder for a VAE is given by:
%
%\begin{equation}
%    P_{\theta} (x|z) = \mathcal{N} (x|f_{\mu_x}(z;\theta), \gamma I)
%\end{equation}
%
%where $\gamma$ is a tuneable scalar hyperparameter---which is typically set as 1 to represent multivariate unit Gaussian distribution---and $I$ is the identity matrix.
%
%We set $\gamma$ as a trainable parameter, as a high $\gamma$ is proven to be responsible for blurry images generated by VAEs, which was often misunderstood as a practical limitation of VAEs~\cite{dai2019diagnosing}.

%$P_\theta (x|z)$
For modeling the unknown PDF of latent variables $P(z|x)$ from which to run the decoder $f_\theta(z)$, we need an encoder function $g(x; \phi)$ which can take an $x$ and return a distribution for $z$ of values that are likely to produce $x$.
%
%This $g$ is modeled with a PDF $Q(z|X)$, and is called a probabilistic encoder. %, which is given by:
%
%\begin{equation}
%    Q_{\phi} (z|x) = \mathcal{N} (z | g_{\mu_z} (x; \phi), g_{{\sigma_z}^2}(x; \phi))
%\end{equation}
%
%where $\phi$ is a set of encoder parameters and $g$ is an encoder function approximated by a deep neural network.
%
This $g$ is called a probabilistic encoder and is modeled with a PDF $Q(z|X)$.
It is designed to produce two outputs, mean $g_{\mu_z}$ and variance $g_{{\sigma_z}^2}$, of the encoded $z$.
Thus, $g$ encodes each $x \in X$ as a distribution, 
where the mean $g_{\mu_z}$ has the highest probability of being reconstructed to $x$.

Since $P(z|x)$ was assumed to be a multivariate Gaussian, the posterior distribution $Q_\phi (z|x)$ 
must {\it match} the $P(z|x)$ so that we can relate $P(x)$ to $\mathbb{E}_{z \sim Q} P(x|z)$, the expected value of generated input $x$ given a latent variable $z$ when $z$ is sampled from the space encoded by $Q$.
This is achieved by minimizing the VAE loss:
$\mathcal{L}(\theta,\phi) = \int_{\mathcal{X}} - \mathbb{E}_{Q_\phi (z|x)}
\lbrack \log P_\theta (x|z) \rbrack + \mathbb{KL} \lbrack Q_\phi (z|x) || P(z) \rbrack \mu_{gt} (dx)$,
where $\mu_{gt}(dx)$ is the ground-truth probability mass of a $dx$ on $X$, which leads to $\int_{X} \mu_{gt}(dx) = 1$.
%Define the ground-truth probability measure $\mu_{gt}(dx)$ as the probability mass of a dx on $X$, where $\int_{X} \mu_{gt}(dx) = 1$.
%
%Given the parameters $\theta, \phi$, the loss function for a simple VAE is 
%
The first term is responsible for minimizing the reconstruction error between $x$ and $\hat{x}$, 
and the second term for making the posterior distribution of $z$ match the prior distribution.
%(Gaussian) 
Interested readers are referred to~\cite{doersch2016tutorial} for a VAE tutorial.

As an example, Figure~\ref{fig:vae} illustrates the structure and the operation of a VAE, with the size of the manifold dimension $\kappa$ set to 2 for a 2D visualization.
The graph on the left illustrates a probabilistic encoder implemented as a neural network, and the graph on the right illustrates a probabilistic decoder implemented as another neural network.
The learnt manifold is illustrated with a cluster of points in a 2D plane that are encoded from the training data, centered around Gaussian mean $(0, 0)$ with a variance of $I_2$.
This space is continuous, smooth, and locally resembles Euclidean space.
Similar images are placed close to each other, as can be seen from the clear separation among inputs from different classes.

For any unseen input $x$, the encoder will produce a manifold representation $z$ that, when passed to the decoder, produces a synthetic image $\hat{x}$ similar to the original image $x$.
This encoder-decoder pair is attached back-to-back during training, but can subsequently be taken apart and used separately for different purposes. 

\section{Test Adequacy}

Coverage criteria are used as proxy measures of software test adequacy.
Structural coverage criteria---such as statement or branch coverage---are frequently used, as they can be measured on the source code, often the only software artifact available.
Although their efficacy and correlation to fault-finding effectiveness have been disputed for long~\cite{groce2014coverage}, they are widely adopted in the industry as they offer a straight-forward and inexpensive gauge of test suite goodness.
In this spirit, several coverage criteria have been proposed to account for the structure of deep neural networks, such as neuron coverage~\cite{pei2017deepxplore}, and their efficacy is also debated~\cite{li2019structural}.
%
%The claimed efficacy and utility of these criteria, however, had been questioned by Li~\etal~\cite{li2019structural} through a series of experiments, showing that no strong correlation can be found between structural coverage and fault-finding effectiveness---catching adversarial inputs, or finding more natural error-revealing inputs.
%
%A more fundamental problem, however, is that the coverage scores may fail to indicate the degree of completeness in testing, a property we take for granted in structural criteria for traditional testing.
%It is not yet clear whether this phenomenon is due to the limitation of the coverage criteria studied so far, or an inherent limitation of the structural coverage criterion itself, but the research in traditional software testing raises a suspicion that it could be the latter~\cite{staats2012danger,groce2014coverage}.

Alternatively, test adequacy can also be measured on other artifacts such as requirements or behavior models.
%~\cite{whalen2006coverage}
%
These offer implementation-independent measures of test adequacy and a further benefit of traceability of tests to higher-level artifacts.
%
%Keeping with this view, we view {\it manifold as a model of the requirements implicit in 
%the training dataset}, as it captures the high-dimensional inputs in a compact representation.
%
One can view the manifold as a compact model of the input space described by the training data.
Thus, coverage criteria defined on the manifold are in effect model-based coverage criteria for ML systems.
%It is then a natural consequence to apply model-based software engineering techniques

\subsection{Manifold Combination Coverage}

%As an illustrative example, the circle-shaped cluster of points in the center of Figure~\ref{fig:vae} shows how the MNIST dataset can be modeled by a VAE with $\kappa=2$, yielding a two-dimensional manifold with a bivariate normal distribution.
%
VAE assumes that each latent dimension is distributed independently.
During training, each latent variable learns some independent semantic feature, although no control can be assumed on what human-understandable feature it learns.
If we conveniently regard a latent variable in a VAE trained on MNIST dataset as encoding the thickness of the hand-written digits, this variable will display a gradation of thickness as it changes its value from the left side to the right side of the Gaussian PDF.
By discretizing this continuous variable to representative categories, such as three levels of thickness, one can easily measure if a test suite covers every category.

The key idea of manifold combination coverage is to apply category partitioning on the latent variables that comprise a manifold, and cover them combinatorially.
%~\cite{nie2011survey}
Combination over the latent dimensions mandates a test suite to include inputs that correspond to every possible interactions among the latent features.
Based on this intuition, we define $k$-section manifold combination coverage as follows; a $t$-way combination coverage can also be defined similarly:
\begin{definition}
{\bf k-section manifold combination coverage:} For a manifold of $\kappa$ latent variables, divide the range of each variable into $k$ sections of equal probability mass.
A test suite $T$ achieves $k$-section manifold combination coverage if there exists one or more sample $z = g_\phi(x). x \in T$ for each of the $k^\kappa$ combinations of sections.
\end{definition}

\subsection{Experiment}

We performed a preliminary experiment to compare the utility of the manifold combination coverage (MCC) with neuron coverage (NC) and neuron boundary coverage (NBC)~\cite{ma2018deepguage} in measuring the diversity of a test dataset.
We first present the coverage scores achieved by each criterion on different subsets of the EMNIST~\cite{cohen2017emnist} dataset which consists of 28,000 images per class.
Next, we quantify the association of coverage obligations---such as a combination of manifold section, or a neuron---to the class label, which is used as a representative semantic feature for this study.
We use Cramer's V, a measure between 0 (no association) and 1 (complete association) of association between two nomial variables, to see how strongly the coverage obligations align with the semantic feature(s) of interest.

For learning the MNIST~\cite{lecun1998mnist} manifold, we trained a VAE of $\kappa=8$.
During encoding, we only took the means from the encoder and discarded the standard deviations to obtain a deterministic representation.
For an instantiation of the manifold combination coverage, we set $k=3$, which yields $3^8 = 6,561$ combinations.
%
%We acknowledge that 
The choice of $\kappa$ and $k$ here was arbitrary, and further study is needed to understand the effect of these choices.

Table~\ref{tbl:coverage} shows the coverage achieved for each subsuite (one per label) of the EMNIST dataset and the MNIST train set (last column).
As known, high NC scores were achieved easily, failing to show any differences among the subsuites.
The Cramer's V score for NC was 0.13, indicating a very low association.
NBC showed low coverage scores, but was able to pronounce some differences between the subsuites.
The V score was 0.42, showing a medium association.
MCC scored 14\% to 30\% per subsuite, and the coverage on the whole set was similar to the coverage achieved by the MNIST training dataset, indicating that MCC was a good measure of completeness.
The V score was 0.83, indicating a high association.

\begin{table}
\centering
\footnotesize
\caption{Coverage achieved by subsets of EMNIST dataset (\%)}
\label{tbl:coverage}
\setlength\tabcolsep{3pt}
\begin{tabular}{|l|r|r|r|r|r|r|r|r|r|r|r|r|} 
\hline
\multicolumn{1}{|c|}{\multirow{2}{*}{Crit.}} & \multicolumn{10}{c|}{Sub-suite by class label}  & \multicolumn{1}{c|}{\multirow{2}{*}{All}} & \multicolumn{1}{c|}{\multirow{2}{*}{Train}}  \\ 
\cline{2-11}
\multicolumn{1}{|c|}{}  & \multicolumn{1}{c|}{0} & \multicolumn{1}{c|}{1} & \multicolumn{1}{c|}{2} & \multicolumn{1}{c|}{3} & \multicolumn{1}{c|}{4} & \multicolumn{1}{c|}{5} & \multicolumn{1}{c|}{6} & \multicolumn{1}{c|}{7} & \multicolumn{1}{c|}{8} & \multicolumn{1}{c|}{9} & \multicolumn{1}{c|}{}  & \multicolumn{1}{c|}{} \\ 
\hline
MCC & 19.6 & 14.1 & 24.6 & 19.6 & 29.9 & 17.6 & 20.1 & 22.7 & 19.5 & 26.9 & 83.6 & 88.8\\ 
\hline
NC & 73.4 & 73.3 & 73.4 & 73.4 & 73.4 & 73.4 & 73.4 & 73.4 & 73.4 &  73.4 & 73.5 & 73.5 \\
\hline
NBC & 3.4 & 0.5 & 6.3 & 4.4 & 4.5 & 6.7 & 3.1 & 4.2 & 5.8 & 3.1 & 14.6 & 0.0 \\
\hline
\end{tabular}
\end{table}

\begin{comment}
\begin{table}
\centering
\caption{Measure of association among class labels and coverage obligations. One indicates a complete association.}
\label{tbl:association}
\begin{tabular}{|l|r|r|r|} 
\hline
\multicolumn{1}{|c|}{Model} & \multicolumn{1}{c|}{MC} & \multicolumn{1}{c|}{NC} & \multicolumn{1}{c|}{NBC}  \\ 
\hline
Fully connected & 0.786 &  0.134 & \\ 
\hline
CNN + batch norm & 0.786 & 0.134 & \\
\hline
\end{tabular}
\end{table}
\end{comment}

\section{Test Generation}

\iffalse
\subsection{Conditional VAE}

{\bf We can move this to the test generation section.}

A vanilla VAE can generate images but not labels.
%
Thus, it may be useful for test input generation, but without the labels, much time has to be spent for assigning labels to solve the oracle problem.
%
When implementing VAE, note that the encoder $q_{\phi} (z|x)$ is conditioned solely on the inputs $x$, and similarly, the decoder $p_{\theta} (x|z)$ models $x$ solely based on the latent-variable vector $z$.
%
Conditional VAE implements a conditional variable $c$ in both the encoder and decoder~\cite{larsen2015autoencoding}.
%
This yields the new loss function
%
\begin{multline} 
  \mathcal{L}(\theta,\phi) = \int_{\mathcal{X}} - \mathbb{E}_{q_\phi (z|x)}
  \lbrack \log p_\theta (x|z,c) \rbrack \\
  + \mathbb{KL} \lbrack q_\phi (z|x,c) || P(z|c) \rbrack \mu_{gt} (dx)
  \label{eq:cvae}
\end{multline}
%
Note that the latent variable $P(z)$ is now distributed under $P(z|c)$, a conditional distribution on $c$.
%
Both the encoder and decoder are conditioned on $c$ as well.
%
When training a CVAE for a classification task, we choose the values of $c$ to be the class labels of the dataset.
%
This gives a specific distribution $P(z|c)$ for each class $c$.
%
By sampling $z$ according to $P(z|c)$, we can significantly increase the probability of obtaining a latent-variable vector $z_0$ such that $p_\theta (x, z_0,c)$ is a valid image of class $c$ ~\cite{sohn2015learning}.

\fi

The goal of testing an image classifier is to find {\it faults} in a model which cause discordance between existing conditions and required conditions~\cite{zhang2019machine}.
%
%For image classification, we informally define fault as a misclassification for in-distribution inputs.
%
One way of finding faults is to run lots of test data on the model, check the outputs against the expectation, and identify fault-revealing inputs.
However, constructing such a test dataset can be difficult and expensive due to the high cost of data collection and laborious labeling.
%
%The situation gets only worse as the accuracy of the model under test approaches a near perfection---a model that achieves 99\% validation accuracy would, on average, require one hundred test cases to find a single fault-revealing input.

Test generation attempts to solve this problem by synthesizing test cases---pairs of inputs and expected outputs---that are likely to trigger faults in the model~\cite{zhang2019machine}.
However, generating realistic fault-revealing images is not a trivial task because images reside in a very high dimension.
%---even for a very small gray-scale image of $28 \times 28$ pixels, there can be $256^{28 \times 28}$ possibilities.
%
Most of the existing approaches either generate synthetic adversarial inputs~\cite{pei2017deepxplore}, or rely on domain-specific metamorphic relations---such as changing the brightness of the image while keeping the object recognizable---for generating realistic inputs.
While both of these approaches are useful, they also come with limitations, as 1) the presence of adversarial inputs does not deliver any insight on the performance of the model on naturally occurring inputs, and 2) the diversity of the generated inputs are bound by the metamorphic relations and the power of the tool which implements the translation.
If one can reliably generate a diverse set of realistic inputs that are also fault-revealing, the testing process can be much accelerated since those {\it counter-examples} can highlight the weakness of the model under test.

For achieving this goal, we propose a novel manifold-based test generation framework, giving a concrete form to the idea also proposed by Yoo~\cite{yoo2019sbst}.
The key is to \textit{treat the manifold as a search space} for finding points from which new inputs can be synthesized.
%
%A well-trained VAE decoder is capable of generating a realistic input ${\hat x}$ that follows the same distribution as the training dataset.
%
A search can be applied on the manifold space to find fault-revealing inputs, and further, the regions of the manifold can be related to the performance of the model under test.
%
%In this paper, we only introduce a random sampling based approach for the tight space limitation.
For brevity, we illustrate the approach with random sampling as the search method.
%
%The fitness is also judged upon the output and the {\it sentiment}~\cite{byun2019input} of the model under test; hence, the overall framework of our approach as shown Figure~\ref{fig:structure}.

%Also, we evaluate the {\it fitness} of a sampled latent vector in the manifold by
%
%We design fitness function that maximizes the realism of the generated images while also maximizing the uncertainty of the model under test.
%
%For doing so, we put a VAE and a target model under test in a closed loop to perform a search, with the goal of obtaining {\it realistic}, {\it interesting}, and fault-revealing test cases.

\subsection{Manifold-based Random Test Generation}

\begin{figure}[]
    \centering
      \includegraphics[width=0.82\columnwidth]{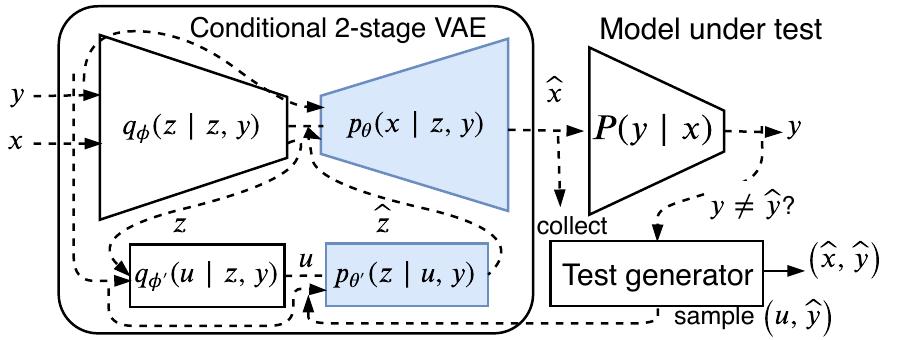}
    \caption{Manifold-based test generator}
    \label{fig:testgen}
\end{figure}
    %
%    New inputs are sampled repeatedly from the manifold, the model under test is scrutinized with each test cases generated, and only the fault-finding inputs are collected in the end.

A vanilla VAE can generate new inputs but not the expected output labels.
To overcome the inefficiency of manually labeling synthesized images, we employ conditional VAE~\cite{larsen2015autoencoding}, which can learn manifold conditioned by class label.
Compared to the vanilla VAE, it has an extra input--the condition $y$ for class label--in the input layer of both the encoder and the decoder.
The loss function used for training is slightly modified VAE loss function, with the PDF of the latent variable $P(z)$ conditioned with $y$ as $P(z|y)$.
%
\begin{comment}
\begin{multline} 
  \mathcal{L}(\theta,\phi) = \int_{\mathcal{X}} - \mathbb{E}_{q_\phi (z|x)}
  \lbrack \log p_\theta (x|z,c) \rbrack \\
  + \mathbb{KL} \lbrack q_\phi (z|x,c) || P(z|c) \rbrack \mu_{gt} (dx)
  \label{eq:cvae}
\end{multline}
\end{comment}
%
When generating a new input, a desired label $y$ needs to be provided to the decoder along with a latent vector $z$.
We also adopt Two-stage VAE~\cite{dai2019diagnosing}, a recent innovation that overcomes blurriness of inputs synthesized by VAE, a problem previously misunderstood as an inherent limitation of the technique.

Figure~\ref{fig:testgen} shows the manifold-based test generation framework.
First of all, a Two-stage VAE is trained one after the other as suggested~\cite{dai2019diagnosing}.
The first VAE is trained to reconstruct the original input image $x$ while learning the first-stage manifold $P(z|x)$.
The second VAE is trained to reconstruct the first-stage encoding $z$ while learning the second-stage manifold $P'(u|z)$.
For both VAEs, the labels of the images are given to the encoder and the decoder as $y$ during training so that we can condition the generated test inputs with desired outputs.
Once the VAEs are trained, the encoders are discarded and only the decoders are used for generating new images.
A new sample $\hat{x}$ can be synthesized by feeding in a choice of vector $u$ with a label $y$ to the second-stage decoder, and in turn $\hat{z}$ and $y$ to the subsequent first-stage decoder.
Each element of the latent vector $u$ is drawn independently from a unit Gaussian distribution, which is the prior assumed on the manifold.
%
%By claiming a control over this second-stage latent space $U$, and by introducing a fitness function, a fitness landscape can be drawn over this space, from which we can generate test inputs of our liking---fault-revealing, {\it realistic}, and {\it interesting}.

\subsection{Experiment}

We performed a preliminary experiment on MNIST dataset to see if the proposed approach can generate realistic fault-revealing test cases.
The model under test was a convolutional neural net with 594,922 trainable parameters which achieved 99.15\% accuracy on the MNIST validation data.
The test generator was a 2-stage conditional VAE with 21,860,515 trainable parameters and $\kappa=32$ latent dimensions, trained for five and half hours on a desktop equipped with a GTX1080-Ti GPU.
Its FID score---which quantifies the realism of the synthesized images---was 6.64 for reconstruction and 8.80 for sampling new inputs, considered very high and comparable to that of the state-of-the-art generative adversarial networks.

\begin{figure}[]
    \centering
    \includegraphics[width=0.92\columnwidth]{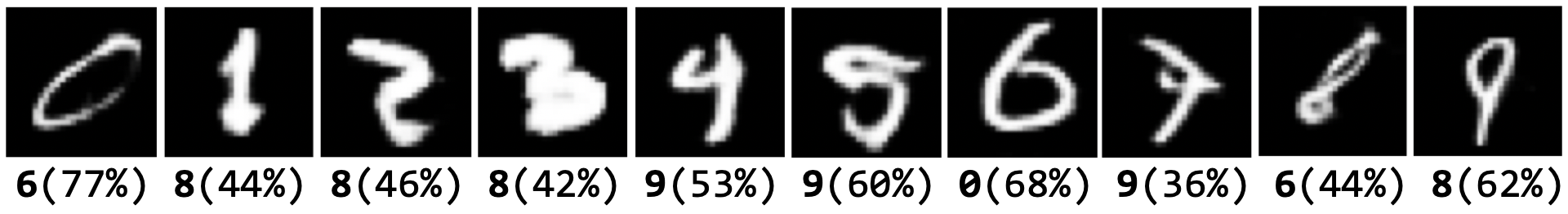}
    \caption{Synthesized fault-revealing test cases}
    \label{fig:faults}
\end{figure}

For obtaining 1,000 fault-revealing test cases on our target, it took about six minutes and 300,256 total test cases (99.67\% accuracy with random testing).
%
%After about six minutes, we could generate 300,256 test cases for obtaining 1,000 fault-revealing ones (99.67\% accuracy with random testing).
%
These test cases were then manually inspected, taking 28 minutes (1.7 sec/test) to filter out marginal cases.
663 inputs were determined to be valid fault-revealing cases.
The rest were either too confusing even to human, or were mislabeled.
A sample of the fault-revealing cases are presented in Figure~\ref{fig:faults} by the order of the true label (automatically generated) with the predicted label and associated probability from the CNN under test.
The results indicate that the fault-revealing inputs are indeed corner cases---peculiar hand-written digits---yet realistic enough to be recognized correctly by human.

\section{Runtime Monitoring}

The need for run-time monitoring arises from the probabilistic nature of the ML systems---even with a near-prefect model, corner cases almost always exist, and architectural mitigation is often the only solution.
Out-of-distribution detection (OOD) attempts to solve this problem by detecting whether a new input falls outside the distribution of the problem domain~\cite{lee2018simple}.
In the spirit of manifold-based ML assurance, we studied an OOD detection method that leverages VAE.
As the VAE learns a distribution which can be handled analytically, the probability of seeing an input $x$ can be computed as a joint probability of encountering a latent vector $z=g_\phi(x)$ in the manifold space.
We defined a function $\mathit{conf}(z) = (\prod_i^{|z|} e^{-0.5 z_i^2})^{1/\kappa}$, which is a joint probability of $z$ under unit normal distributions normalized to a value in the unit interval.
The result can be seen as a confidence measure for an input $x$ being in-distribution.

\begin{figure}[]
    \centering
    \includegraphics[width=0.92\columnwidth]{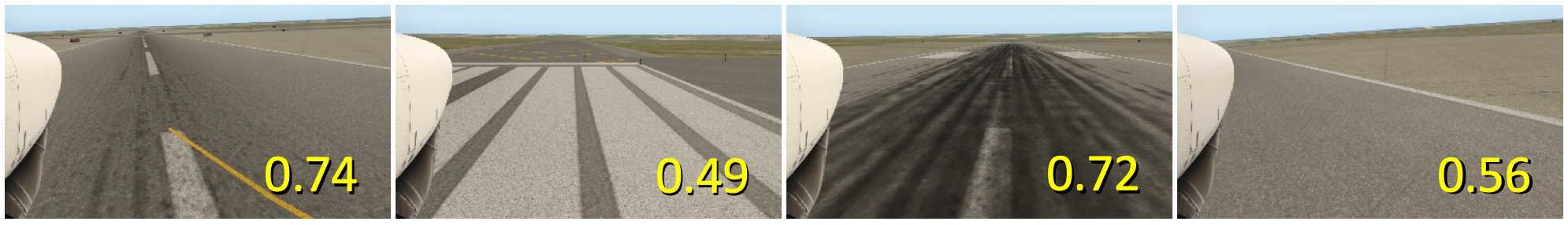}
    \caption{Confidence scores assigned by manifold monitor}
    \label{fig:taxinet}
\end{figure}

We assessed the feasibility of this approach by integrating it into an autonomous aircraft taxiing system prototype.
In such systems, detection must be performed in real-time so that appropriate actions can be taken as needed.
The VAE-based monitor was able to provide real-time confidence estimate with a moderate computational overhead, and showed that it can flag corner-case inputs.
Figure~\ref{fig:taxinet} illustrates some exemplary cases in 1) normal condition, 2) end of the runway, 3) runway region with skid marks, and 4) edge of the runway.
The confidence score roughly corresponded with the probability mass of each case in the training data.

\section{Future Work}
We proposed novel approaches supported by early results to leverage manifolds learned by VAEs for quality assurance of ML systems.
%This paper proposed novel approaches of using manifold for quality assurance of ML systems.
%
%We introduced three manifold-based assurance techniques---test adequacy measurement, test generation, and runtime monitoring---and presented promising preliminary experimental results for those.
%
Further work is required to explore and fully exploit its potential.
To keep $t$-way combinatorial testing tractable, small $t$ values are desirable.
Such an approach and its effectiveness need rigorous investigation.
%{\bf coverage:} 
The efficacy of manifold coverage is likely to be enhanced when combined with measures that are
designed to prioritize corner-cases, such as surprise adequacy or uncertainty~\cite{kim2019guiding,byun2019input}.
This idea needs to be developed further and evaluated.
The idea of searching the manifold space for generating {\it interesting} test inputs needs further development.
Such approaches need to be empirically compared with each other and random generation.
The use of VAE as a runtime monitor to detect potential problem behaviors needs further refinement.
Specifically, ways to assess confidence and evaluate the metrics' performance are needed.
%For runtime monitoring, we need to establish a good metric for evaluating the confidence we produce with VAE monitor.
%
%Overall, {\bf a broad last statement}.
Finally, the dimensionality reduction achieved by this approach could benefit formal analysis techniques that may not be otherwise applicable at scale.
Whether VAEs provide appropriate abstractions for formal verification needs investigation.

\begin{comment}

This result, however, does not imply that MCC is better than other criteria for other purposes as well.
%
A measure designed specifically to credit corner cases---such as NBC or Bayesian uncertainty~\cite{byun2019input}---may work better at achieving a high fault-finding.
%
Even if that is the case, gauging the completeness of testing is important for establishing confidence, which is what manifold-based criteria is shown to be good at.
%
A further investigation is necessary for assessing the effectiveness of these criteria, or combinations of them, in various settings for various purposes.
\end{comment}

%\section{Related Work}
%\input{sections/related.tex}

%\section{Conclusion and Future Work}
%\input{sections/conclusion.tex}

%%
%% The acknowledgments section is defined using the "acks" environment
%% (and NOT an unnumbered section). This ensures the proper
%% identification of the section in the article metadata, and the
%% consistent spelling of the heading.
\begin{acks}
This work was supported by AFRL and DARPA under contract FA8750-18-C-0099.
\end{acks}

%%
%% The next two lines define the bibliography style to be used, and
%% the bibliography file.
\bibliographystyle{ACM-Reference-Format}
\bibliography{vaetestgen}

\end{document}